\let\svthefootnote\thefootnote

\documentclass{research4cacm}
\usepackage{array,supertabular}
\usepackage{amsmath,amssymb,amsfonts,mathrsfs}
\usepackage{url,microtype,relsize}
\newcommand{\ac}[1]{#1}
\newcommand{\acsfont}[1]{#1}

\newcommand{\algname}[1]{{\small\texttt{#1}}}

\newcommand\RepackingStations{\ensuremath{S_{\text{bidding}}}}

\newcommand{\citeauthor}[1]{\cite{#1}} %
\newcommand{\emcite}[1]{\cite{#1}} %
\newcommand{\eg}{e.g.,\ }	%

\begin{document}
\CopyrightYear{2017} %

\title{Deep Optimization for Spectrum Repacking
\thanks{This paper builds in part on a conference publication by the same authors, ``Solving the Station Repacking Problem'', which was published at AAAI 2016.}}
\numberofauthors{3} %
\author{
\alignauthor
Neil Newman\\
       \affaddr{Department of Computer Science}\\
       \affaddr{University of British Columbia, Canada}\\
       \email{newmanne@cs.ubc.ca}
\alignauthor
Alexandre Fr\'echette\\
\affaddr{Department of Computer Science}\\
\affaddr{University of British Columbia, Canada}\\
\email{afrechet@cs.ubc.ca}
\alignauthor Kevin Leyton-Brown\\
       \affaddr{Department of Computer Science}\\
       \affaddr{University of British Columbia, Canada}\\
       \email{kevinlb@cs.ubc.ca}
}

\maketitle
\begin{abstract}
Over 13 months in 2016--17 the US Federal Communications Commission conducted an ``incentive auction'' to repurpose radio spectrum from broadcast television to wireless internet. In the end, 
the auction yielded \$19.8 billion, 
\$10.05 billion
of which was paid to 175 broadcasters for voluntarily relinquishing their licenses across 14 UHF channels.
Stations that continued broadcasting were assigned potentially new channels to fit as densely as possible into the channels that remained. The government netted more than \$7 billion (used to pay down the national debt) after covering %
costs. A crucial element of the auction design was the construction of a solver, dubbed SATFC, that determined whether sets of stations could be ``repacked'' in this way; it needed to run every time a station was given a price quote. This paper describes the process by which we built SATFC. We adopted an approach we dub ``deep optimization'', taking a data-driven, highly parametric, and computationally intensive approach to solver design. More specifically, to build SATFC we designed software that could pair both complete and local-search SAT-encoded feasibility checking with a wide range of domain-specific techniques, such as constraint graph decomposition and novel caching mechanisms that allow for reuse of partial solutions from related, solved problems. We then used automatic algorithm configuration techniques to construct a portfolio of eight complementary algorithms to be run in parallel, aiming to achieve good performance on instances that arose in proprietary auction simulations. To evaluate the impact of our solver in this paper, we built an open-source reverse auction simulator. We found that within the short time budget required in practice, SATFC solved more than 95\% of the problems it encountered. Furthermore, the incentive auction paired with SATFC produced nearly optimal allocations in a restricted setting and substantially outperformed other alternatives at national scale.
\end{abstract}

\section{Introduction}
\label{sec:Introduction}
\let\thefootnote\svthefootnote
Many devices, including broadcast television receivers and cell phones, rely on the transmission of electromagnetic signals. These signals can interfere with each other, so transmission is regulated: e.g., in the US, by the Federal Communications Commission (\ac{FCC}). Since electromagnetic spectrum suitable for wireless transmission is a scarce resource and since it is difficult for a central authority to assess the relative merits of competing claims on it, since 1994 the \ac{FCC} has used \emph{spectrum auctions} to allocate broadcast rights (see, e.g., \cite{milgrom2004putting}). Many regulators around the world have followed suit. At this point, in the US (as in many other countries), most useful radio spectrum has been allocated.
Interest has thus grown in the \emph{reallocation} of radio spectrum from less to more valuable uses. Spectrum currently allocated to broadcast television has received particular attention, for two reasons. First, over-the-air television has been losing popularity with the rise of cable, satellite, and streaming services. 
Second, the upper \ac{UHF} frequencies used by TV broadcasters are particularly well suited to wireless data transmission on mobile devices---for which demand is growing rapidly---as they can penetrate walls and travel long distances \cite{knutson_verizon_2014}. 

It thus made sense for at least some broadcasters to sell their licenses to wireless internet providers willing to pay for them. Ideally, these trades would have occurred bilaterally and without government involvement, as occurs in many other markets. However, two key obstacles made such trade unlikely to produce useful, large-scale spectrum reallocation, both stemming from the fact that wireless internet services require large, contiguous blocks of spectrum to work efficiently. First, a buyer's decision about which block of spectrum to buy would limit the buyer to trading only with broadcasters holding licenses to parts of that block; it could be hard or impossible to find such a block in which all broadcasters were willing to trade. Second, each of these broadcasters would have ``holdout power'', meaning the broadcaster could demand an exorbitant payment in exchange for allowing the deal to proceed. The likely result would have been very little trade, even if broadcasters valued the spectrum much less than potential buyers.

A 2012 Act of Congress implemented a clever solution to this problem. It guaranteed each broadcaster interference-free coverage in its broadcast area on \emph{some} channel, but not necessarily on its currently used channel. This meant that if a broadcaster was unwilling to sell its license it could instead be moved to another channel, solving the holdout problem. To free up the channel that would permit this move to take place, broadcast rights could be bought from another station in the appropriate geographical area, even if this second station did not hold a license for spectrum due for reallocation. In what follows, we call such an interference-free reassignment of channels to stations a \emph{feasible repacking}.

These trades and channel reassignments were coordinated via a novel spectrum auction run by the \ac{FCC} between March 2016 and April 2017, dubbed the \emph{Incentive Auction}. It consisted of two interrelated parts. The first was a \emph{forward auction} that sold large blocks of upper UHF spectrum to interested buyers in a manner similar to previous auctions of unallocated spectrum. The key innovation was the second part: a \emph{reverse auction} that was specially designed to perform well in the Incentive Auction \cite{milgrom2014deferred,li2015obviously}.
It identified both a set of broadcasters who would voluntarily give up their broadcast rights and prices at which they would be compensated, simultaneously ensuring that all remaining broadcasters could feasibly be repacked in the unsold spectrum. 
The choice of how much spectrum to reallocate, called the \emph{clearing target}, linked these two parts: the incentive auction alternated between reverse and forward auction stages with progressively shrinking clearing targets until revenue generated by the forward auction covered the cost of purchasing and reassigning stations in the reverse auction.

We now describe the reverse auction's rules in more detail.
First, all participating stations are given initial price quotes and respond either that they agree to sell their broadcast rights at the quoted price or that they ``exit the auction'' (decline to participate), meaning that they will be guaranteed some interference-free channel. The auction then repeatedly iterates over the active bidders. Every time a bidder $i$ is considered, the software first checks whether $i$ can be feasibly repacked along with all exited stations. If such a feasible repacking exists, $i$ is given a (geometrically) lower price quote and again has the options of accepting or exiting. Otherwise, $i$ is \emph{frozen}: its price stops descending and it is no longer active. The auction ends when all bidders are either frozen, exited, or receive price quotes of zero. 

The problem of checking the feasibility of repackings is central to the reverse auction, likely to arise tens of thousands of times in a single auction. Unfortunately, this problem is NP-complete, generalizing graph coloring. The silver lining is that interference constraints were known in advance---they were derived based on the locations and broadcast powers of existing television antennas---and so it was reasonable to hope for a heuristic algorithm that achieved good performance on the sorts of problems that would arise in a real auction. However, identifying an algorithm that would be fast and reliable enough to use in practice remained challenging. Since each feasibility check depends on the results of those that came before---if a station is found to be frozen, it cannot exit---these problems must be solved sequentially. Time constraints for the auction as a whole required that the auction iterate through the stations at least twice a day, which worked out to a time cutoff on the order of minutes. It was thus inevitable that some problems would remain unsolved. Luckily, the auction design is robust to such failures, treating them as proofs of infeasibility at the expense of raising the cost required to clear spectrum.

This paper describes our experience building \ac{SATFC} 2.3.1, the feasibility checker used in the reverse auction. We leveraged automatic algorithm configuration approaches to derive a portfolio of complementary algorithms that differ in their underlying (local and complete) search strategies, \ac{SAT} encodings, constraint graph decompositions, domain-specific heuristics, and use of a novel caching scheme. 
We use the term ``deep optimization'' to refer to this approach,%
	\footnote{There exists a large body of prior work that investigates the use of algorithm configuration to design novel algorithms from large, parameterized spaces (some of which, indeed, we have coauthored); we believe, however, that the work described in this paper is the most consequential application of such techniques to date. Much of the literature just mentioned focuses on algorithm configuration tools \cite{hutter2009paramils,irace,HutHooLey11-smac,gga,Hydra-2010,isac} (which we take as given in this paper) rather than algorithm design methodology. Most work in the latter vein either addresses the much broader problem of algorithm synthesis (e.g., \cite{aeon,easysyn,westfold2001}) or defines the overall approach only implicitly (e.g., \cite{khudabukhsh2016satenstein}). The most prominent exception is ``programming by optimization'' \cite{pbo}; however, it emphasizes connections to software engineering and does not limit itself to parametric design spaces.}
with the goal of emphasizing its conceptual similarity to deep learning.
Classical machine learning relied on features crafted based on expert insight, model families selected manually, and model hyperparameters tuned essentially by hand. Deep learning has shown that it is often possible to achieve substantially better performance by relying less on expert knowledge and more on enormous amounts of computation and huge training sets. Specifically, deep learning considers parametric models of very high dimension, using expert knowledge only to identify appropriate invariances and model biases, such as convolutional structure. (In some cases it is critical that these models be ``deep'' in the sense of having long chains of dependencies between parameters, but in other cases great flexibility can be achieved even with models only a couple of levels deep; e.g., \cite{ZhangBHRV16}.) We argue that a similar dynamic applies in the case of heuristic algorithms for discrete optimization, which aim to achieve good performance on some given dataset rather than in the worst case. Traditionally, experts have designed such heuristic algorithms by hand, iteratively conducting small experiments to refine their designs. We advocate an approach in which a computationally intensive procedure is used to search a high-dimensional space of parameterized algorithm designs to optimize performance over a large set of training data. We aim to minimize the role played by expert knowledge, restricting it to the identification of parameters that could potentially lead to fruitful algorithm designs. We also encourage deep dependencies via chains of parameters each of whose meaning depends on the value taken by one or more parents. 

Overall, this paper demonstrates the value of the deep optimization approach via the enormous performance gains it yielded on the challenging and socially important problem of spectrum repacking. After formally stating the station repacking problem, we define our large algorithm design space and the search techniques we used. We assess the results on problems that arose in runs of our new open-source reverse auction simulator, investigating both our solver's runtime and its impact on economic outcomes.

\section{The Station Repacking Problem}
\label{sec:Problem}

We now describe the station repacking problem in more detail.\footnote{
	Similar problems have been studied in other contexts, falling under the umbrella of \emph{frequency assignment problems}. See \eg \emcite{aardal2007models} for a survey and a discussion of applications to mobile telephony, radio and TV broadcasting, satellite communication, wireless LANs, and military operations. We are not aware of other published work that aims to optimize feasibility checking in the Incentive Auction setting.}
Each television station in the US and Canada $s \in \mathcal{S}$ is currently assigned to a channel $c_s \in \mathcal{C} \subseteq \mathbb{N}$ that ensures that it will not excessively interfere with other, nearby stations. (Although Canadian stations did not participate in the auction, they were eligible to be reassigned new channels.) The FCC determined pairs of channel assignments that would cause harmful interference based on a complex, grid-based physical simulation (``OET-69'' \cite{oet69}); this pairwise constraint data is publicly available \cite{constraintfiles}. %
Let $\mathcal{I} \subseteq (\mathcal{S}\times\mathcal{C})^2$ denote a set of \emph{forbidden station--channel pairs} $\{(s,c),(s',c')\}$, each representing the proposition that stations $s$ and $s'$ may not concurrently be assigned to channels $c$ and $c'$, respectively. The effect of the auction was to remove some broadcasters from the airwaves and to reassign channels to the remaining stations from a reduced set. This reduced set was defined by a \emph{clearing target}, fixed for each stage of the reverse auction, corresponding to some channel $\overline{c} \in \mathcal{C}$ such that all stations are only eligible to be assigned channels from $\overline{\mathcal{C}} = \{c \in \mathcal{C} \mid c < \overline{c}\}$. Each station can only be assigned a channel on a subset of $\mathcal{\overline{C}}$, given by a \emph{domain} function $\mathcal{D}: \mathcal{S} \to 2^{\overline{\mathcal{C}}}$ that maps from stations to these reduced sets. The \emph{station repacking problem} is then the task of finding a repacking $\gamma : \mathcal{S} \to \overline{\mathcal{C}}$ that assigns each station a channel from its domain that satisfies the interference constraints: i.e., for which $\gamma(s) \in \mathcal{D}(s)$ for all $s\in S$, and $\gamma(s) = c \Rightarrow \gamma(s') \not= c'$ for all $\{(s,c),(s',c')\} \in \mathcal{I}$. A problem instance thus corresponds to a set of stations $S\subseteq \mathcal{S}$ and channels $C\subseteq \overline{\mathcal{C}}$ into which they must be packed, with domains $\mathcal{D}$ and constraints $\mathcal{I}$ implicitly being restricted to $S$ and $C$; we call the resulting restrictions $D$ and $I$.

\begin{figure}[t]
	\centering
	\includegraphics[width=.9\columnwidth]{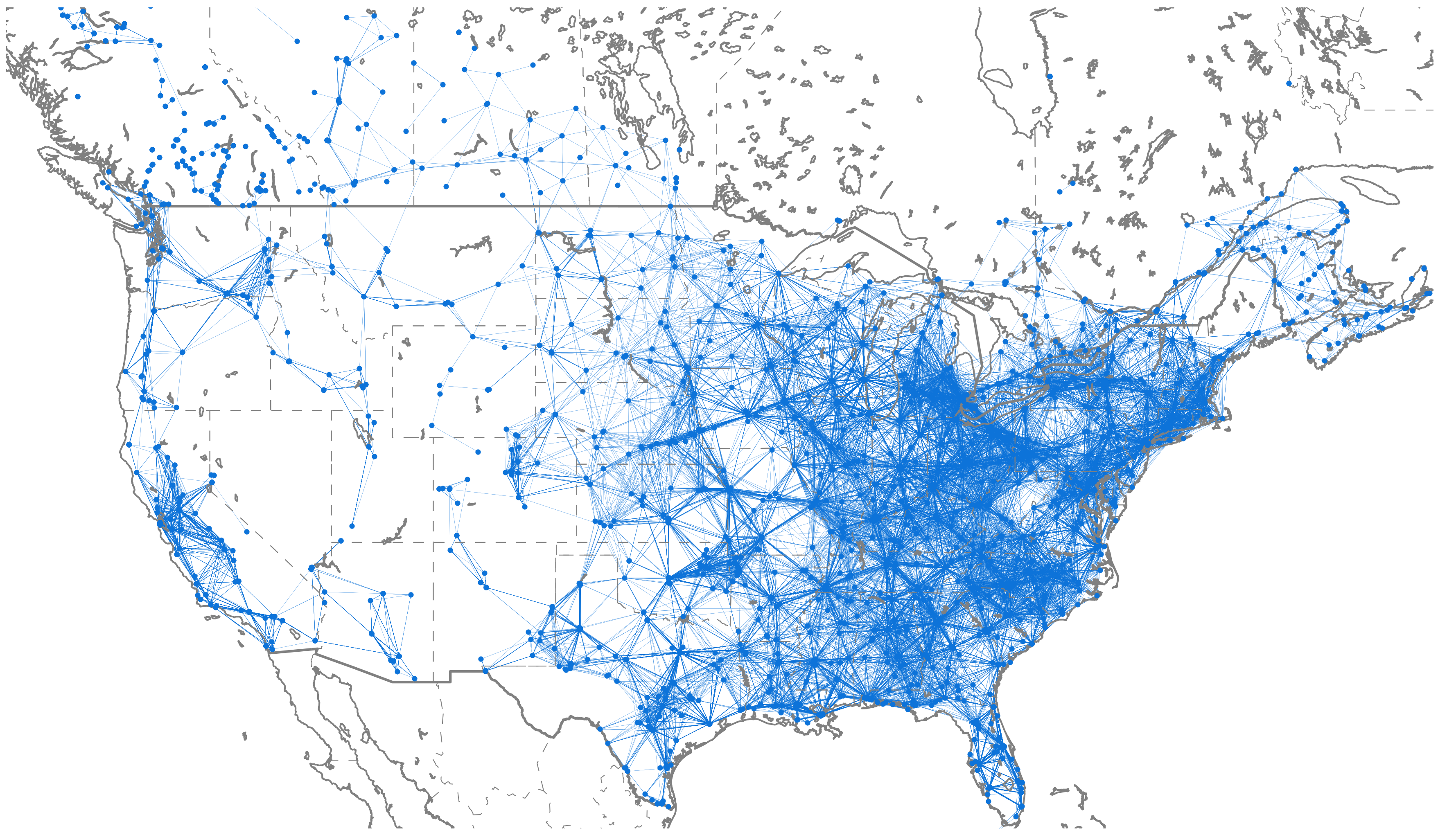}
	\caption{Interference graph visualizing the \ac{FCC}'s %
		constraint data \protect\cite{constraintfiles} (2\,990 stations; 2\,575\,466 channel-specific interference constraints).}
	\label{fig:interference-graph}
\end{figure}

Why should we hope that this (\ac{NP}-complete) problem can be solved effectively in practice? First, we only need to be concerned with problems involving subsets of a fixed set of stations and a fixed set of interference constraints: those describing the television stations currently broadcasting in the United States and Canada. 
Let us define the \emph{interference graph} as an undirected graph in which there is one vertex per station and an edge exists between two vertices $s$ and $s'$ if the corresponding stations participate together in any interference constraint: i.e., if there exist $c,c' \in C$ such that $\{(s,c),(s',c')\} \in I$. Figure~\ref{fig:interference-graph} shows the Incentive Auction interference graph. As it turns out, interference constraints come in two kinds. \emph{Co-channel constraints} specify that two stations may not be assigned to the same channel; \emph{adjacent-channel constraints} specify that two stations may not be assigned to two nearby channels. Hence, any forbidden station--channel pairs are of the form $\{(s,c),(s',c + i)\}$ for some stations $s,s'\in \mathcal{S}$, channel $c\in\mathcal{C}$, and $i \in \left\{0,1,2\right\}$. Furthermore, channels can be partitioned into three equivalence classes: LVHF (channels 1--6), HVHF (channels 7--13), and UHF (channels 14--51) with the property that no interference constraint involves channels in more than one band.

Second, note that we are not interested in optimizing worst-case performance even given our fixed interference graph, but rather in achieving good performance on the sort of instances generated by actual reverse auctions. These instances depend on the order in which stations exit the auction, which depends on stations' valuations, which depend in turn (among many other factors) on the size and character of the population reached by their broadcasts. The distribution over repacking problems is hence far from uniform. 

Third, descending clock auctions repeatedly generate station repacking problems by adding a single station $s^+$ to a set $S^-$ of provably repackable stations. This means that every station repacking problem $(S^-\cup\{s^+\},C)$ comes with a partial assignment $\gamma^-: S^- \to C$ that we know is feasible on restricted station set $S^-$; we will see in what follows that this fact is extremely useful.

Finally, many repacking problems are trivial: in our experience, problems involving only VHF channels can all be solved quickly; furthermore, the vast majority of UHF problems can be solved greedily simply by checking whether $s^+$ can be augmented directly with $\gamma^-$. However, solving the remaining problems is crucial to the economic outcomes achieved by the auction (as we show in Section~\ref{sec:ComparingOutcomes}). In what follows, we restrict ourselves to ``non-trivial'' UHF problems that cannot be solved by greedy feasibility checking. 

\section{A Deep Optimization Approach}

As we show in our experiments (see Section~\ref{sec:RuntimePerformance}), off-the-shelf solvers could not solve a large enough fraction of station repacking problems to be effective in practice. 
To do better, we needed a customized algorithm optimized to perform well on our particular distribution of station repacking problems. We built our algorithm via the deep optimization approach, meaning that we aimed to use our own insight only to identify design ideas that showed promise, relegating the work of combining these ideas and evaluating the performance of the resulting algorithm on realistic data (see Section~\ref{sec:data}) to an automatic search procedure.

\subsection{The Design Space}

Our first task was thus to identify a space of algorithm designs to consider. This was not just a pen-and-paper exercise, since each point in the space needed to correspond to runnable code. 
We focused on encoding station repacking as a propositional satisfiability (SAT) problem. The SAT formalism is well suited to station repacking, which is a pure feasibility problem with only combinatorial constraints. (It may also be possible to achieve good performance with MIP or other encodings; we did not investigate such alternatives in depth.) The SAT reduction is straightforward: given a station repacking problem $(S,C)$ with domains $D$ and interference constraints $I$, we create a Boolean variable $x_{s,c} \in \{\top,\bot\}$ for every station--channel pair $(s,c) \in S\times C$, representing the proposition that station $s$ is assigned to channel $c$. We then create three kinds of clauses: (1) $\bigvee_{d\in D(s)} x_{s,d} \;\forall s \in S$ (each station is assigned at least one channel); (2) $\neg{x_{s,c}} \vee \neg{x_{s,c'}} \;\forall s \in S,\,\forall c,c'\neq c\in D(s)$ (each station is assigned at most one channel); (3) $\neg{x_{s,c}} \vee \neg{x_{s',c'}}\;\forall \{(s,c),(s',c')\} \in I$ (interference constraints are respected). Note that (2) is optional: if a station is assigned more than one channel, we can simply pick one channel to assign it from among these channels arbitrarily. We thus created a parameter indicating whether to include these constraints. In the end, a SAT encoding of a problem involving all stations at a clearing target of 36 involved 73\,187 variables and 2\,917\,866 clauses. %

\subsubsection{Selecting Solvers}

Perhaps the most important top-level parameter determines which SAT solver to run. (Of course, each such solver will have its own (deep) parameter space; other parameters will describe design dimensions orthogonal to the choice of solver, as we will discuss in what follows.) The SAT community has developed a very wide variety of solvers and made them publicly available (see e.g., \emcite{jarvisalo2012international}). In principle, we would have made it possible to choose every solver that offered even reasonable performance. However, doing so would have been too costly from the perspective of software integration and (especially) reliability testing. We thus conducted initial algorithm configuration experiments (see Section~\ref{sec:configuration}) on 20 state-of-the-art SAT solvers, drawn mainly from SAT solver competition entries collected in AClib \cite{hutter2014aclib}. We illustrate the performance of their default configurations later in Figure~\ref{fig:ecdf_defaults}; most improved at least somewhat from their default configurations as a result of algorithm configuration. We identified two solvers that ended up with the strongest post-configuration performance---one complete and one based on local search---both of which have been shown in the literature to adapt well to a wide range of SAT domains via large and flexible parameter spaces. Our first solver was \algname{clasp} \cite{gebser2007clasp}, an open-source solver based on conflict-driven nogood learning (98 parameters). Our second was the open-source SATenstein framework \cite{khudabukhsh2016satenstein}, which allows arbitrary composition of design elements taken from a wide range of high-performance stochastic local search solvers (90 parameters). %

\subsubsection{Using the Previous Solution}
\label{sec:prevsol}

While adapting \algname{clasp} and \algname{SATenstein} to station repacking data yielded substantial performance improvements, neither reached a point sufficient for deployment in the real auction. To do better, it was necessary to leverage specific properties of the incentive auction problem. Rather than committing to specific speedups, we exposed a wide variety of possibilities via further parameters.
We began by considering two methods for taking advantage of the existence of a partial assignment $\gamma^-$. 
The first method checks whether a simple transformation of $\gamma^-$ is enough to yield a satisfiable repacking. Specifically, we construct a small SAT problem in which the stations to be repacked are $s^+$ and all stations $\Gamma(s^+)\subseteq S$ neighboring $s^+$ in the interference graph, fixing all other stations $S \setminus \Gamma(s^+)$ to their assignments in $\gamma^-$. Any solution to this reduced problem must be a feasible repacking; however, if the reduced problem is infeasible we cannot conclude anything. However (depending on the value of a parameter), we can keep searching: unfixing all stations that neighbor a station in $\Gamma(s^+)$, and so on. 

Our second method uses $\gamma^-$ to initialize local search solvers. Such solvers search a space of complete variable assignments, typically following gradients to minimize an objective function such as the number of violated constraints, with occasional random steps. They are thus sensitive to their starting points. Optionally, we can start at the assignment given by $\gamma^-$ (randomly initializing variables pertaining to $s^+$). We can also optionally redo this initialization on some fraction of random restarts.

\subsubsection{Problem Simplification}
\label{sec:problemsimplification}

Next, we considered three preprocessing techniques that can simplify station repacking problems. First, we added the option to run the {arc consistency} algorithm, repeatedly pruning values from each station's domain that are incompatible with every channel on a neighboring station's domain. 

Second, we enabled elimination of unconstrained stations. A station $s$ is unconstrained if, given any feasible assignment of all of the other stations in $S \setminus s$, there always exists some way of feasibly repacking $s$.  Unconstrained stations can be removed without changing a problem's satisfiability status. Various algorithms exist for identifying unconstrained stations; we determine this choice via a parameter. (All such stations can be found via a reduction to the polytime problem of eliminating variables in a binary CSP \cite{Cohen20151127}; various sound but incomplete heuristics run more quickly but identify progressively fewer unconstrained stations.)

Third, the interference graph induced by a problem may consist of multiple connected components; we can optionally run a linear-time procedure to separate them into distinct SAT problems. We only need to solve the component to which $s^+$ belongs: $\gamma^-$ supplies feasible assignments for all others. Arc consistency and unconstrained station removal can simplify the interference graph by removing edges and nodes respectively. This can shrink the size of the component containing $s^+$ and make this technique even more effective.

\subsubsection{Containment Caching}
\label{sec:cache}
Finally, we know that every repacking problem will be derived from a restriction of the interference graph to some subset of $\mathcal{S}$. We know this graph in advance of the auction; this suggests the possibility of doing offline work to precompute solutions. However, our graph has 2\,990 nodes, and the number of restricted graphs is thus $2^{2990} \approx 10^{900}$. Thus, it is not possible to consider all of them offline.

Not every restricted problem is equally likely to arise in practice. To target likely problems, we could simply run a large number of simulations and cache the solution to every repacking problem encountered. Unfortunately, we found that it was extremely rare for problems to repeat across sufficiently different simulator inputs, even after running hundreds of simulations (generating millions of instances and costing years of CPU time). However, we can do better than simply looking for previous solutions to a given repacking problem. If we know that $S$ is repackable then we know the same is true for every $S' \subseteq S$ (and indeed, we know the packing itself---the packing for $S$ restricted to the stations in $S'$). Similarly, if we know that $S$ was not packable then we know the same for every $S' \supseteq S$. This observation dramatically magnifies the usefulness of each cached entry $S$, because each $S$ can be used to answer queries about an exponential number of subsets or supersets. This is especially useful because sometimes it can be harder to find a repacking for subsets of $S$ than it can be to find a repacking for $S$.

We call a cache meant to be used in this way a \emph{containment cache}, because it is queried to determine whether one set contains another (i.e., whether the query contains the cache item or vice versa). To the best of our knowledge, containment caching is a novel idea. A likely reason why this scheme is not already common is that querying a containment cache is nontrivial: one cannot simply index entries with a hash function; instead, an exponential number of keys can match a given query. We were nevertheless able to construct an algorithm that solved this problem quickly in our setting. We observe that containment caching is applicable to any family of feasibility testing problems generated as subsets of a master set of constraints, not just to spectrum repacking.

In more detail, we maintain two caches, a \emph{feasible cache} and an \emph{infeasible cache}, and store each problem we solve in the appropriate cache. We leverage the methods from Section~\ref{sec:problemsimplification} to enhance the efficiency of our cache, storing full instances for \ac{SAT} problems and the smallest simplified component for \ac{UNSAT} problems. When asked whether it is possible to repack station set $S$, we first check whether a subset of $S$ belongs to the infeasible cache (in which case the original problem is infeasible); if we find no matches, we decompose the problem into its smallest simplified component and check if the feasible cache contains a superset of those stations, in which case the original problem is feasible.

\subsection{Searching the Design Space}
\label{sec:configuration}

Overall, our design space had 191 parameters, nested as much as 4 levels deep. We now describe how we searched this space to building a customized solver.
Identifying a set of parameters that optimize a given algorithm's performance on a given dataset is called \emph{algorithm configuration}. There exist a wide variety of algorithm configuration tools \cite{hutter2009paramils,irace,HutHooLey11-smac,gga}. We used Sequential Model-based Algorithm Configuration (SMAC) \cite{HutHooLey11-smac}, the publicly available method that arguably achieves the best performance (see \eg \cite{cssc2014}). SMAC uses the ``Bayesian optimization'' approach of interleaving random sampling and the exploration of algorithm designs that appear promising based on a learned model.

Unfortunately, even after performing algorithm configuration, it is rare to find a single algorithm that outperforms all others on instances of an \ac{NP}-complete problem such as \ac{SAT}. This inherent variability across solvers can be exploited by \emph{algorithm portfolios} \cite{gomes2001algorithm,Satzilla03,SATzilla-Full}. Most straightforwardly, one selects a small set of algorithms with complementary performance on problems of interest and, when asked to solve a new instance, executes them in parallel. 
\label{sec:SATFC}
Of course, we wanted to construct such algorithm portfolios automatically as part of our deep optimization approach. We did this by using a method called \algname{Hydra} \cite{Hydra-2010} which runs iteratively, at each step directing the algorithm configurator to optimize marginal gains over the given portfolio. This allows \algname{Hydra} to find algorithms that may perform poorly overall but that complement the existing portfolio. Overall, we ran \algname{Hydra} for eight steps, thereby producing a portfolio of novel solvers (dubbed \ac{SATFC}) that could run on a standard eight-core workstation. The Incentive Auction used \ac{SATFC} 2.3.1, which is available online at \url{https://github.com/FCC/SATFC}.

\section{Data from Auction Simulations}
\label{sec:data}
During the development of SATFC \cite{Frechette:2016:SSR:3015812.3015917} the FCC shared with us a wide range of anonymized problem instances that arose in auction simulations they performed in order to validate the auction design. These formed the ``training set'' we used in the deep optimization process when constructing SATFC 2.3.1. These simulations explored a very narrow set of answers to the questions of which stations would participate and how bidders would interact with the auction mechanism; they do not represent a statement either by us or by the \ac{FCC} about how these questions were resolved in the real auction (indeed, by law the answers will not be revealed to us or to the public for two years). It is of course impossible to guarantee that variations in the assumptions would not have yielded computationally different problems. 

While SATFC 2.3.1 is itself open-source software, it is unfortunately impossible for us to share the data that was used to build it.
In this paper, we have opted for what we hope is the next best thing: \emph{evaluating} SATFC 2.3.1 and various alternatives using a publicly available test set. We thus wrote our own reverse auction simulator and released it as open source software (see \url{http://cs.ubc.ca/labs/beta/Projects/SATFC}). 
We used this simulator to simulate 20 auctions, in each case randomly sampling bidder valuations from a publicly available model \cite{doraszelski2016ownership} using parameters obtained directly from its authors. This model specifies stations' values for broadcasting in UHF, $v_{s,\text{UHF}}$. Of course, a station has no value for going off air: $v_{s,\text{OFF}} = 0$. In some cases the reverse auction can reassign a UHF station to a channel in one of two less valuable VHF bands (LVHF, HVHF) in exchange for lesser compensation. We assume that $v_{s,\text{HVHF}} = \frac{2}{3}v_{s,\text{UHF}}$ and $v_{s,\text{LVHF}} = \frac{1}{3}v_{s,\text{UHF}}$. We excluded from our simulator all stations for which the authors of the model were unable to supply us with parameters: stations outside the mainland US and Hawaii, all US VHF stations, 
and an additional 25 US UHF stations. This left us with 1\,638 eligible US stations. We further included all Canadian stations in our simulations: because the auction rules forbade them from being paid to leave their home bands, we did not need to model their valuations. Specifically, from Canada we included 113 \ac{LVHF} stations, 332 \ac{HVHF} stations, and 348 \ac{UHF} stations. 

We set the auction's opening prices to the values announced by the \ac{FCC} in November 2015 \cite{stationinfofiles}. We assumed that stations chose to participate in the reverse auction if their opening price offer for going off air was greater than their valuation for remaining on air in their current band. We assumed that stations always selected the option that myopically maximized their utility. 
We used the interference constraints and station domains announced by the \ac{FCC} in November 2015 \cite{constraintfiles}. For each simulation, we used the largest clearing target for which we could find a feasible assignment for the non-participating stations; in all cases this led to a clearing target of 84 Mhz, corresponding to a maximum allowable channel of 36. We note that this is the amount of spectrum actually cleared by the Incentive Auction. 
Just like the real auction, an auction simulator needs a feasibility checker to determine which price movements are possible. We used \ac{SATFC} 2.3.1 with a cutoff of 60 seconds. %
We sampled 10\,000 ``nontrivial'' UHF problems uniformly at random from all of the problems across all simulations to use as our dataset, where we defined nontrivial problems as those that could not be solved by greedily augmenting the previous solution. Fewer than 3\% of UHF problems in our simulations were nontrivial. This test set consisted of 9\,482 feasible problems, 121 infeasible problems, and 397 problems that timed out at our one minute cutoff and therefore have unknown feasibility. 

\section{Runtime Performance}
\label{sec:RuntimePerformance}

\begin{figure}
	\centering
	\includegraphics[width=\columnwidth]{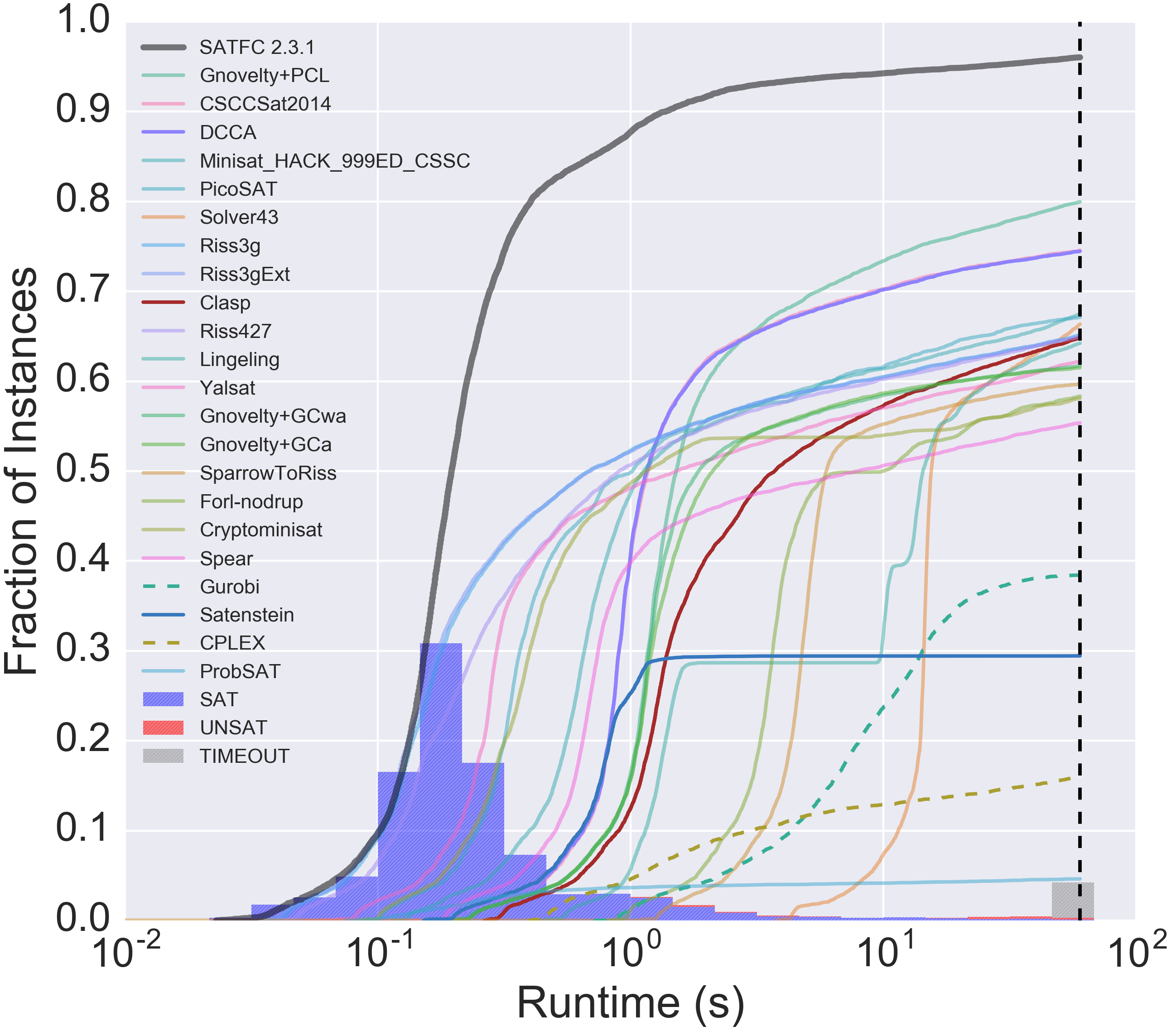}
	\caption{Empirical Cumulative Density Function (\ac{ECDF}) of runtimes for default configurations of \ac{MIP} and \ac{SAT} solvers and for \ac{SATFC} 2.3.1. The curves show fraction of instances solved ($y$ axis) within different amounts of time ($x$ axis; note the log scale). The legend is ordered by percentage of problems solved before the cutoff. The histogram indicates density of \ac{SAT} and \ac{UNSAT} instances binned by their (fastest) runtimes; unsatisfiable instances constituted fewer than 1\% of solved instances.}
	\label{fig:ecdf_defaults}
\end{figure}

We now evaluate SATFC's performance by contrasting it with various off-the-shelf alternatives. The FCC's initial investigations included modeling the station repacking problem as a mixed-integer program (MIP) and using off-the-shelf solvers paired with problem-specific speedups \cite{da143}. 
Unfortunately, the problem-specific elements of this solution were not publicly released, so we cannot evaluate them in this article. Instead, to assess the feasibility of a MIP approach, we ran what are arguably the two best-performing MIP solvers---CPLEX and Gurobi---on our test set of 10\,000 non-trivial instances. To encode the station repacking problem as a MIP, we created a variable $x_{s,c} \in \{0,1\}$ for every station--channel pair, representing the proposition that station $s$ is assigned to channel $c$. We imposed the constraints $\sum_{c \in D(s)} x_{s,c} = 1\ \forall\ s \in S$ and $x_{s,c} + x_{s',c'} \leq 1\ \forall\ \{(s,c),(s',c')\} \in I$, ensuring that each station is assigned to exactly one channel and that interference constraints are not violated. Both MIP solvers solved under half of the instances within our cutoff time of one minute; the results are shown in Figure~\ref{fig:ecdf_defaults}. Such performance would likely have been insufficient for deployment in practice, since it implies unnecessarily high payments to many stations. %

As already discussed, there exist a wide variety of SAT solvers that are available for use off the shelf. Figure~\ref{fig:ecdf_defaults} illustrates the performance of the 20 state-of-the-art solvers we considered in our initial configuration experiments in their default configurations. With few exceptions, the \ac{SAT} solvers outperformed the \ac{MIP} solvers, as can be seen by comparing the solid and dashed lines in Figure~\ref{fig:ecdf_defaults}. However, runtimes and percentages of instances solved by the cutoff time were still not good enough for us to recommend deployment of any of these solvers in the actual auction. The best solver in its default configuration, \algname{Gnovelty+PCL}, was able to solve the largest number of problems---79.96\%---within the cutoff. (As mentioned earlier, the SATenstein design space includes \algname{Gnovelty+PCL} alongside many other solvers.)
The parallel portfolio of all 20 solvers from Figure~\ref{fig:ecdf_defaults} was little better, being able to solve only 81.58\% of problems. 

We now turn to \ac{SATFC} 2.3.1. This 8-solver parallel portfolio stochastically dominated every individual solver that we considered and achieved very substantial gains after a few tenths of a second. It solved 87.73\% of the problems in under a second and 96.03\% within the one-minute cutoff time. The histogram at the bottom of the figure indicates satisfiability status of instances solved by SATFC grouped by runtime; our instances were overwhelmingly satisfiable. 

\label{sec:Experiments}

\section{Impact on Economic Outcomes}\label{sec:ComparingOutcomes}

\begin{figure*}[t]
	\includegraphics[width=.98\columnwidth]{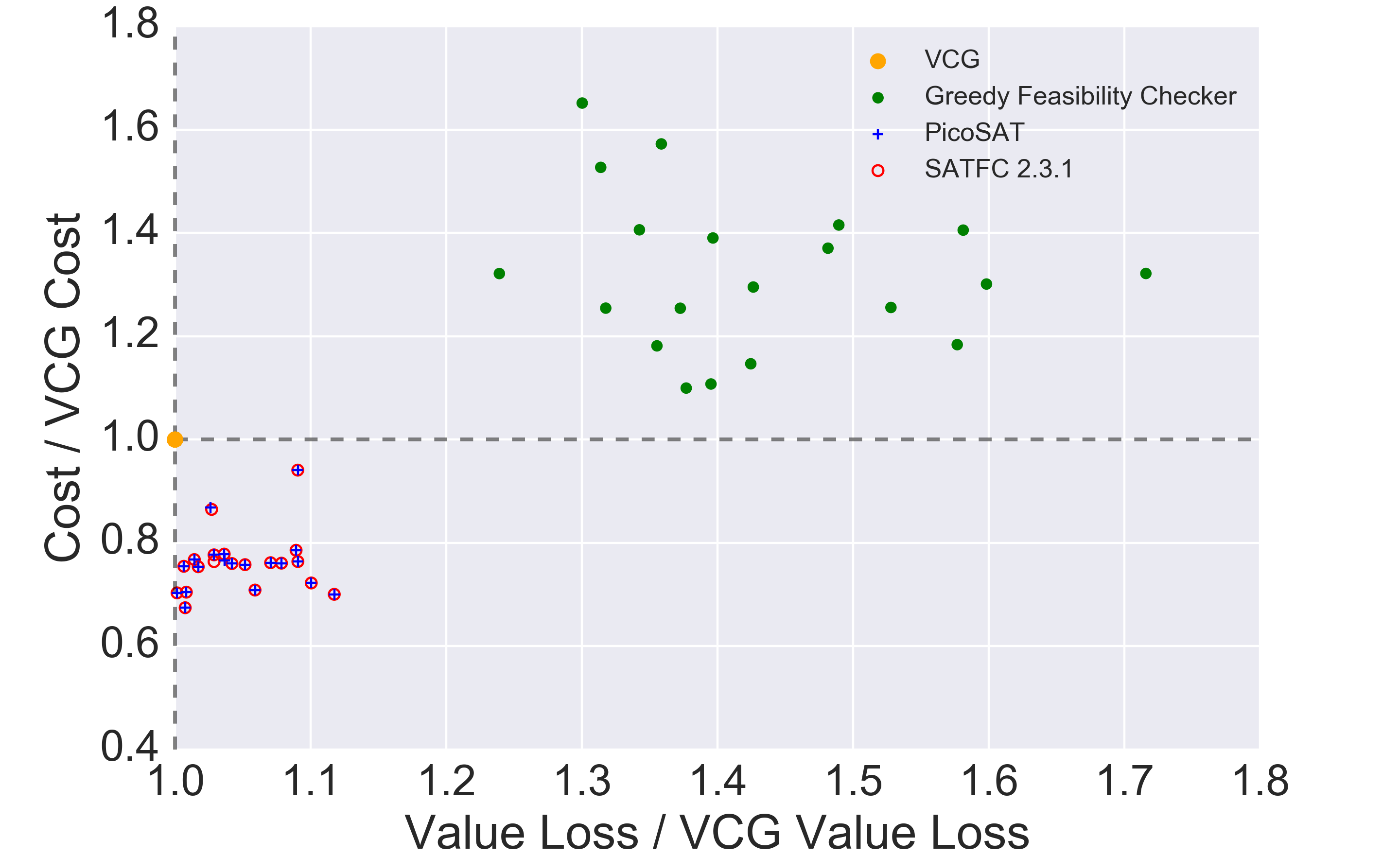}\hfill
	\includegraphics[width=.98\columnwidth]{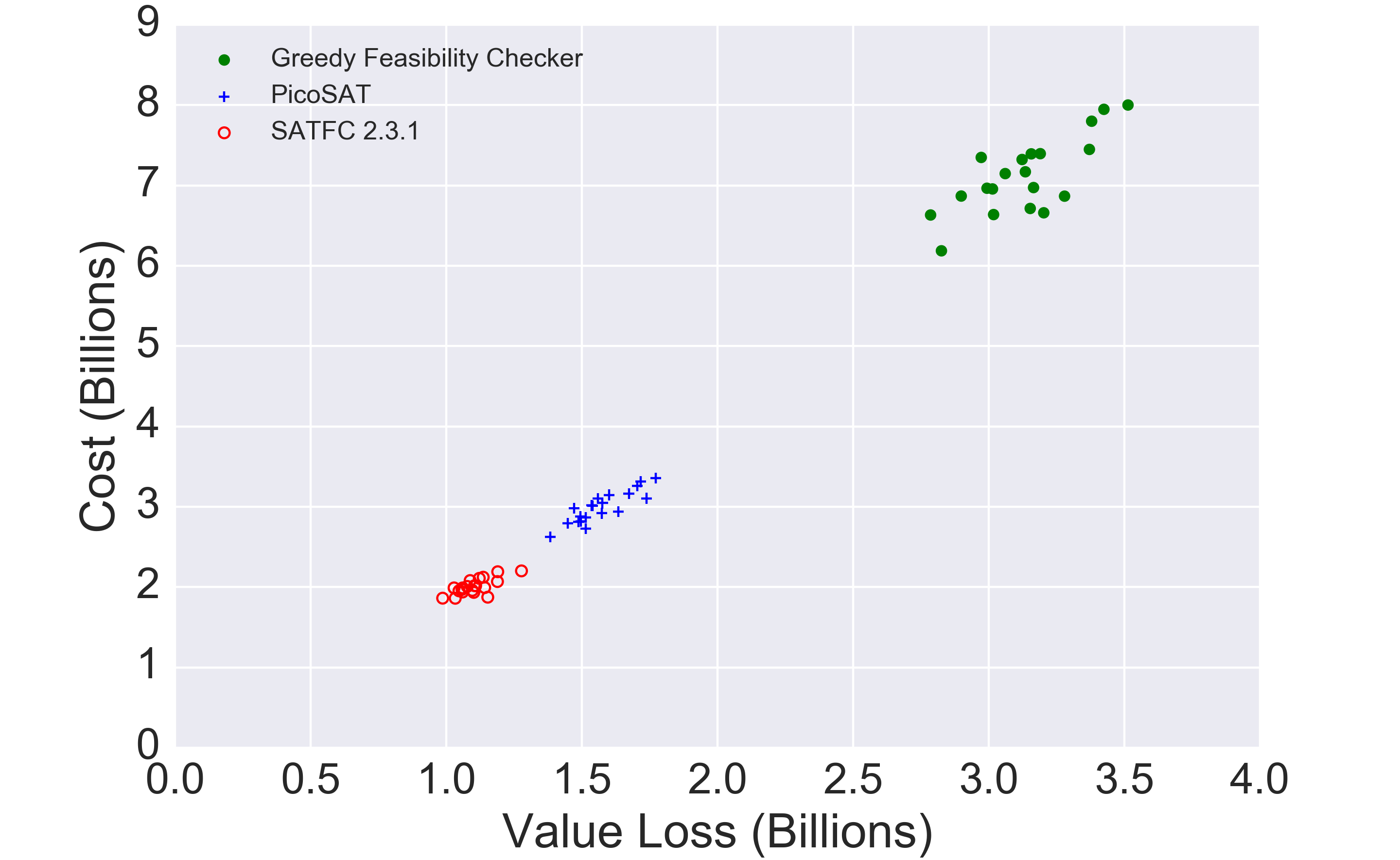}
	\caption{Comparing value loss and cost of the greedy feasibility checker, \algname{picoSAT}, and \ac{SATFC} 2.3.1 for 20 different value profiles. \emph{(Left)} Fraction of \ac{VCG} cost versus fraction of \ac{VCG} value loss for Greater New York City simulations.  All \ac{VCG} points lie at (1,1). \emph{(Right)} Value loss and cost for national simulations.}
	\label{fig:vcg}
\end{figure*}

We now ask whether SATFC's improved performance is likely to have translated into a better economic outcome in the Incentive Auction, assessing both cost and efficiency. 
The cost of an auction is the sum of payments to the winning stations. 
To assess efficiency we measured the total value \emph{lost} by the auction, comparing the sums of values of stations for their allocated bands both before and after the auction.\footnote{
	The more standard measure of efficiency---the sum of stations' values for their allocated bands---has the same optimum. Value loss has the advantage that it is influenced only by stations that a feasibility checker is unable to repack in their home bands; it is thus more appropriate for comparing feasibility checkers. The more standard measure is sensitive to changes in the values of easy-to-repack stations, even those that do not participate in any interference constraints. }
If we can find an efficient repacking $\gamma^*$, then we can compute the additional fraction of value lost by some other repacking $\gamma$. We call this the \emph{value loss ratio}: $\frac{\sum_{s \in \mathcal{S}} v_{s,\text{pre}\left(s\right)} - v_{s,\text{post}\left(\gamma, s\right)}}{\sum_{s \in \mathcal{S}} v_{s,\text{pre}\left(s\right)} - v_{s,\text{post}\left(\gamma^*, s\right)}}$,
where $\text{pre}(s)$ returns the band to which $s$ was assigned before the auction and $\text{post}(\gamma,s)$ returns either the band to which $s$ is assigned under channel assignment $\gamma$ or \acsfont{OFF} if $s$ is not assigned to a band under $\gamma$. 
When it is intractable to compute $\gamma^*$, we resort to comparing the absolute value loss between different assignments.

Given our interest in the efficiency of the reverse auction, it is natural to compare it to the \ac{VCG} mechanism, which always chooses the optimal packing $\gamma^*$. \ac{VCG} pays losing stations nothing and pays each winning station $s$ the difference between the sum of values of stations other than $s$ for $\gamma^*$ and the sum of the same stations' values for a packing that is optimal subject to the constraint that $s$ does not win. We identified these optimal packings using the MIP encoding from Section~\ref{sec:RuntimePerformance} with two changes. First, we added the objective of maximizing the aggregate values of the participating stations: maximize $\sum_{s \in \RepackingStations} \sum_{c \in D\left(s\right)} x_{s,c} \cdot v_{s,\text{band}\left(c\right)}$, where  $\text{band}\left(c\right)$ is a function that returns the corresponding band for a given channel. Second, we allowed the option of not assigning a channel to a bidding station.

\subsection{Greater New York City Simulations}

Unfortunately, it was impossible to solve these optimization problems at a national scale, even given several days of computing time. We therefore constructed tractable problems by restricting ourselves to stations in the vicinity of New York City, which we chose because it corresponds to one of the most densely connected regions in the interference graph. More specifically, we dropped all Canadian stations and restricted ourselves to the \ac{UHF} band using the smallest possible clearing target (maximum allowable channel of 29). Using the interference graph induced by these restrictions, we then dropped every station whose shortest path length to a station in New York City exceeded two. The result was a setting with 218 stations and 78\,499 channel-specific interference constraints, yielding a MIP encoding with 2\,465 variables and 78\,717 constraints.

We randomly generated 20 different valuation profiles, using the methodology described in Section~\ref{sec:data} but restricting ourselves to the stations in the restricted interference graph. For each valuation profile, we conducted four simulations. The first was of a VCG auction; we computed allocations and payments using \algname{CPLEX}, solving all \ac{MIP}s optimally to within $10^{-6}$ absolute \ac{MIP} gap tolerance.  We also ran three reverse auction simulations for each valuation profile, varying the feasibility checker to consider two alternatives to \ac{SATFC}. The first is the greedy feasibility checker, which represents the simplest reasonable feasibility checker and thus serves as a baseline. The second is the default configuration of \algname{picoSAT}. To our knowledge, alongside MIP approaches this is the only other solver that has been used in publications on the Incentive Auction \cite{DBLP:journals/corr/KearnsD14,cramton2015design}, probably because we showed it to be the best among a set of alternatives in an early talk on the subject \cite{kevintalk}.

In total, our 80 simulations consumed over 5 years of CPU time (dominated by the VCG simulations). 
Figure~\ref{fig:vcg} (left) illustrates the results. Each point shows the value loss ($x$ axis) and cost ($y$ axis) of a single simulation; in both cases, these quantities are normalized by the corresponding quantity achieved by VCG for the same valuation profile. The \ac{SATFC} simulations had a mean value loss ratio of 1.048 and a mean cost ratio of 0.760, indicating that the reverse auction achieved nearly optimal efficiency at much lower cost than VCG. The \algname{picoSAT} results were nearly identical, differing in only two value profiles, and then only slightly. 
Both the \ac{SATFC} and \algname{picoSAT} runs dominated the \algname{greedy} runs according to both metrics; on average, reverse auctions based on \algname{greedy} cost 1.742 times more and lost 1.366 times as much broadcaster value than those based on \ac{SATFC}. Despite these differences, all of the solvers were able to solve a very large fraction of the feasibility checking problems encountered in their respective simulations (which took different trajectories once two solvers differed in their ability to solve a given problem): 99.978\%, 99.945\%, and 99.118\% for \ac{SATFC}, \algname{picoSAT}, and \algname{greedy} respectively (including trivial problems).

\subsection{National Simulations}

We were more interested in economic outcomes at the national scale, even though we could not simulate \ac{VCG} in such a large setting. We generated 20 valuation profiles for our full set of stations and ran reverse auction simulations using our three feasibility checkers. In total, these experiments consumed over 5 days of CPU time.

All solvers were again able to solve a large fraction of the problems they encountered: 99.902\%, 99.765\%, and 98.031\% for \ac{SATFC}, \algname{picoSAT}, and \algname{greedy} respectively (including trivial problems). The economic impact of these differences is illustrated in Figure~\ref{fig:vcg} (right). In this graph, $x$- and $y$-axis values correspond to unnormalized value loss and cost respectively. Each \ac{SATFC} and \algname{picoSAT} simulation again dominated its greedy counterpart in both efficiency and cost. Averaging over all of our observations, reverse auctions based on greedy feasibility checking cost 3.550 times (\$5.114 billion) more and lost 2.850 times as much (\$2.030 billion) broadcaster value than those based on \ac{SATFC}. At the larger scale we also found that \ac{SATFC} dominated \algname{picoSAT} in every simulation: on average, \algname{picoSAT} auctions cost 1.495 times (\$987 million) more and lost 1.427 times as much (\$469 million) value than \ac{SATFC} auctions. 

\section{Conclusions}
\label{sec:Conclusions}

Station repacking in the Incentive Auction is a difficult but important problem, with progress translating into significant gains in both government expenditures and social welfare. We designed a customized solution to this problem using an approach we dub deep optimization. Specifically, we drew on a large parameterized design space to construct a strong portfolio of heuristic algorithms: \ac{SATFC} 2.3.1, an open-source solver that was used in the real auction. To evaluate it for this paper, we conducted experiments with a new reverse auction simulator. We found that replacing \ac{SATFC} with an off-the-shelf feasibility checker resulted in both efficiency losses and increased costs. It thus appears likely that our efforts led to significant economic benefits to broadcasters, the US government, and the American public.

\section{Acknowledgments}

\smaller
We gratefully acknowledge support from Auctionomics and the FCC; valuable conversations with Paul Milgrom, Ilya Segal, and James Wright; help from Peter West in conducting some of the experiments reported in Section 6; contributions (mostly in the form of code) from past research assistants Nick Arnosti, Guillaume Saulnier-Comte, Ricky Chen, Alim Virani, Chris Cameron, Emily Chen, Paul Cernek; experimental infrastructure assistance from Steve Ramage; and help gathering data from Ulrich Gall, Rory Molinari, Karla Hoffman, Brett Tarnutzer, Sasha Javid, and others at the FCC. This work was funded by Auctionomics and by NSERC via a Discovery Grant and an E.W.R.\ Steacie Fellowship; it was conducted in part at the Simons Institute for Theoretical Computer Science at UC Berkeley.

\bibliographystyle{abbrv} %
\bibliography{satfc}  
\balancecolumns
\end{document}